\documentclass{new_tlp_updated}
\usepackage{mathptmx}

\usepackage{times,helvet,courier}
\usepackage{amsfonts,amsmath,amssymb}
\usepackage{xspace,epsfig}
\usepackage{booktabs}
\usepackage{longtable}
\usepackage{multirow}
\usepackage{algorithm}
\usepackage[noend]{algorithmic}
\usepackage{lscape}
\usepackage{url}
\usepackage{setspace}
\usepackage{rotating}
\usepackage[justification=centering]{caption}

\def\ba{\begin{array}}
\def\ea{\end{array}}
\def\beq{\begin{equation}}
\def\eeq#1{\label{#1}\end{equation}}
\def\beqq{\begin{equation*}}
\def\eeqq{\end{equation*}}

\def\ii#1{\hbox{\it #1\/}}
\def\smti{SMTI\xspace}
\def\smt{SMT\xspace}
\def\sm{SM\xspace}

\def\clingo{{\sc Clingo}\xspace}
\def\cmodels{{\sc Cmodels}\xspace}
\def\zchaff{{\sc ZChaff}\xspace}
\def\lingeling{{\sc Lingeling}\xspace}

\def\lar{\leftarrow}

\newcommand{\algrule}[1][.3pt]{\par\vskip.3\baselineskip\hrule height #1\par\vskip.3\baselineskip}

\title[Stable Marriage Problems: ASP, SAT, ILP, CP, Local Search]{Stable Marriage Problems with Ties and Incomplete Preferences: An Empirical Comparison of ASP, SAT, ILP, CP, and Local Search Methods}

\author[Eyupoglu et al.]
{SELIN EYUPOGLU$^1$, MUGE FIDAN$^1$, YAVUZ GULESEN$^1$, ILAYDA BEGUM IZCI$^1$,
\and
BERKAN TEBER$^1$, BATURAY YILMAZ$^1$, AHMET ALKAN$^2$, ESRA ERDEM$^1$ \\
$^1$ Computer Science, Faculty of Engineering and Natural Sciences, Sabanci University, Istanbul, Turkey \\
$^2$ Economics, Faculty of Arts and Social Sciences, Sabanci University, Istanbul, Turkey \\
}

\begin{document}
\thispagestyle{plain}

\label{firstpage}

\maketitle

\begin{abstract}
We study a variation of the Stable Marriage problem, where every man and every woman express their preferences as preference lists which may be incomplete and contain ties. This problem is called the Stable Marriage problem with Ties and Incomplete preferences (\smti). We consider three optimization variants of \smti,  Max Cardinality, Sex-Equal and Egalitarian, and empirically compare the following methods to solve them: Answer Set Programming, Constraint Programming, Integer Linear Programming. For Max Cardinality, we compare these methods with Local Search methods as well. We also empirically compare Answer Set Programming with Propositional Satisfiability, for \smti instances.

\end{abstract}


\section{Introduction}\label{sec:intro}

Matching problems have been studied in economics, starting with the seminal paper of \citeN{gale1962college}, which has led to a Nobel Prize in 2012, utilizing game theory methods with the goal of a mechanism design. Matching problems are about markets where individuals are matched with individuals, firms, or items, typically across two sides, as in employment~\cite{roth1990twosided} (e.g., who works at which job), kidney donation (e.g., who receives which transplantable organ)~\cite{manlove2014paired,roth2005pairwise}, and marriages~\cite{manlove1999stable,gale1962college} (e.g., who marries with whom). In each problem, preferences of individuals, firms, or items are given, possibly along with other information (e.g., the quotas of the universities in university entrance)~\cite{alkan2003mathematical,alkan2003gale}.

One of the well-known matching problems is the Stable Marriage Problem (\sm). In \sm, for a set of $n$ men and $n$ women, we are given the preferences of individuals: for each man, a complete ranking of the women is specified as preferred partners; similarly, for each woman, a complete ranking of the men is specified as preferred partners. The goal is to marry all men and women (i.e., to find $n$ couples) in such a way that marriages are stable: no man and woman in different couples prefer each other to their partners.

We consider a variant of \sm, called \smti, where rankings may be incomplete (i.e., some partners are not acceptable) or may include ties (i.e., some partners are preferred equally). We investigate three hard variants of \smti~\cite{kato1993complexity,manlove2002hard}, that aim to compute optimal stable matchings with respect to different measures of fairness: sex-equality (preferences of men and women are considered to be equally important), egalitarian (preferences of every individual are considered to be equally important), maximum cardinality (minimizes the number of singles).

We present a variety of methods to solve these problems, using Answer Set Programming (ASP)~\cite{mar99,nie99,lif02,simons02,BrewkaEL16}, Integer Linear Programming (ILP)~\cite{kantorovich39}, Constraint Programming (CP)~\cite{jaffar1987,VanHentenryck89,rossi2006}, and Local Search (including Hill-Climbing~\cite{linK1973,selman2006} and Genetic Algorithms~\cite{holland92}).

The ASP formulations of \smti and its hard variants (Sex-Equal \smti,  Egalitarian \smti, Max Cardinality \smti) are novel (Section~\ref{sec:asp}); they are implemented for the ASP solver \clingo \cite{GebserKKS19}. We consider the ILP formulation of Max Cardinality \smti, by Kwanashie and Manlove~\citeyear{kwanashie2014} as a basis, and introduce the ILP formulations for Sex-Equal \smti and Egalitarian \smti (Section~\ref{sec:ilp}); they are implemented for Gurobi and Google-OR Tools (MIP and CP).  We consider the local search algorithms introduced by Gelain et al.~\citeyear{gelain2013local} and Haas et al.~\citeyear{haas2020} for Max Cardinality \smti (Section~\ref{sec:local}), and implement them with slight variations.   We compare these methods empirically over a large set of randomly generated instances (Section~\ref{sec:experiments}).

We also compare ASP with propositional satisfiability (SAT)~\cite{Gomes07,biere2009} over randomly generated \smti instances. For a comparison of ASP with SAT, we adapt the ASP implementation of \smti for \cmodels~\cite{cmodels} that utilizes the SAT solver \zchaff~\cite{chaff} to compute solutions. We also adapt the SAT formulation of \smti by \citeN{Drummond15sat}~(SAT-E) to include ties, and use the SAT solver \lingeling~\cite{Biere2010LingelingPP} to compute solutions.

The implementations and the benchmarks are available at {\small \url{https://github.com/KRR-SU/SMTI}}.

\section{The Stable Marriage Problems with Ties and Incomplete Lists (\smti)}\label{sec:smti}

The Stable Marriage problem with Ties and Incomplete lists (\smti) is defined by a set~$M$ of men, a set~$W$ of women, for each man $x \in M$ a partial ordering $\succ_{x}$ over $W_x \subseteq W$, and for each woman $y \in W$ a partial ordering $\succ_{y}$ over $M_y \subseteq M$.

Let $\ii{mrank}: M \times W \mapsto \{1,\dots,|W|\}$ be a partial function such that $\ii{mrank}(x,y) = r$ represents that a woman $y$ is man $x$'s $r$th preferred choice with respect to $\succ_{x}$ and $\ii{wrank}: W \times M \mapsto \{1,\dots,|M|\}$ be a partial function such that $\ii{wrank}(y,x)=r$ represents that a man $x$ is woman $y$'s $r$th preferred choice with respect to $\succ_{y}$.

A man $x$ is {\em acceptable} to a woman $y$ if $\ii{wrank}(x,y)$ is defined. Similarly, a woman~$y$ is acceptable for a man $x$ if $\ii{mrank}(x,y)$ is defined.

A {\em matching} for a given \smti instance is a partial function $\mu: M \mapsto W$. A man $x$ is single if $\mu(x)$ is undefined and a woman $y$ is single if $\mu^{-1}(y)$ is undefined.

A pair $(x,y)$ of a man and a woman is called a {\em blocking pair} for a matching $\mu$ if the following conditions hold:
\begin{itemize}
\item[A1] $x$ and $y$ are acceptable to each other,
\item[A2] $x$ and $y$ are not married to each other (i.e., $\mu(x) \neq y$),
\item[A3]
\begin{itemize}
 \item[(a)] $x$ and $y$ are both single,
    \item[(b)]  $\ii{mrank}(x,y) < \ii{mrank}(x,\mu(x))$ and $y$ is single,
   \item[(c)] $\ii{wrank}(y,x) < \ii{wrank}(y,\mu^{-1}(y))$ and $x$ is single or
\item[(d)] $ \ii{mrank}(x,y) < \ii{mrank}(x,\mu(x))$ and  $ \ii{wrank}(y,x) < \ii{wrank}(y,\mu^{-1}(y))$.
\end{itemize}
\end{itemize}

A matching for \smti is called {\em stable} if it is not blocked by any pair of agents.
Note that we consider weakly stable matchings. It is assumed that marriage to an acceptable partner is preferred over being single.

We consider three hard variants of \smti~\cite{kato1993complexity,manlove2002hard}, that aim to compute optimal stable matchings with respect to different measures of fairness: sex-equality, egalitarian and maximum cardinality.

\paragraph{\bf Egalitarian \smti.}
Egalitarian \smti maximizes the total satisfaction of the preferences of all agents. Since the preferred agents have lower rankings, this total satisfaction is maximized when the sum of ranks of all agents is minimized.

Let $\mu$ be a matching and $\mathcal{M}$ denote the set of matchings in a given problem. For every man $x \in M$, we define the satisfaction $c_\mu(x)$ of $x$'s preferences with respect to $\mu$ as follows: $c_\mu(x) {=} R$ if $\ii{mrank}(x,\mu(x)){=}R$. Similarly, for every woman $y \in W$, we define the satisfaction $c_\mu(y) {=} R$ if $\ii{wrank}(y,\mu^{-1}(y)){=}R$. Then, the total satisfaction of preferences of all agents is defined as follows: $c(\mu) = \sum_{x\in M \cup W} c_\mu(x).$  Then, a matching $\mu {\in} \mathcal{M}$ with the minimum $c(\mu)$ is {\em egalitarian.}

\paragraph{\bf Sex-Equal \smti.}
Sex-Equal \smti maximizes the equality of satisfaction among sexes. We define the sex equality by the following cost function $c(\mu) {=}   \vert \sum_{x \in M}c_\mu(x) - \sum_{y \in W}c_\mu(y) \vert.$ Then, a matching $\mu {\in} \mathcal{M}$ with the minimum $c(\mu)$ is {\em sex-equal.}

\paragraph{\bf Max Cardinality \smti.}
Max Cardinality \smti maximizes the number of matched pairs. A matching  $\mu {\in} \mathcal{M}$ is a {\em maximum cardinality matching} if it is a matching that maximizes $\vert \mu \vert$. The number of matched pairs is maximized when the number of singles is minimized.


\section{Solving \smti Problems using ASP and SAT}
\label{sec:asp}

We formalize input of an \smti instance $\langle M, W, \ii{mrank}, \ii{wrank} \rangle$ in ASP by a set $F_I$ of facts using atoms of the forms $\ii{man}(x)$ (``$x$ is a man in $M$''), $\ii{woman}(y)$ (``$y$ is a woman in $W$''),  $\ii{mrank}(x,y,r)$ (i.e., $\ii{mrank}(x,y) = r$) and  $\ii{wrank}(y,x,r)$ (i.e., $\ii{wrank}(y,x) = r$). In the ASP formulation $P$ of \smti, the variables $x$, $x1$ denote men in $M$ and $y$, $y1$ denote women in $W$.

The first pair of rules of the program~$P$ characterize a set of individuals of the opposite set for each man and woman who they prefer over being single:
$$
\ba l
  \ii {maccept}(x, y) \lar  \ii{mrank}(x, y, r). \\
  \ii {waccept}(y, x) \lar \ii{wrank}(y, x, r). \\
\ea
$$
and the concept of mutual acceptability:
$$
\ba l
  \ii {acceptable}(x, y) \lar \ii{maccept}(x, y), \ii{waccept}(y, x).
\ea
$$
We define preferences of man $x$ (i.e., $x$ prefers  $y$ to $y1$) and woman $y$  (i.e., $y$ prefers $x$ to $x1$) in terms of rankings.
$$
\ba l
  \ii{mprefer}(x, y, y1) \lar  \ii{mrank}(x, y1, r), \ii{mrank}(x, y, r1), r > r1. \\
  \ii{wprefer}(y, x, x1) \lar  \ii{wrank}(y, x1, r), \ii{wrank}(y, x, r1),  r > r1. \\
\ea
$$
We define a matching between men and women where both parties find each other acceptable with the cardinal constraint of 1 for each man.
$$
\ba l
 $\{$ \ii{marry}(x,y) : \ii{acceptable}(x,y)$\}1$ \lar \ii{man}(x).\\
\ea
$$
In order to guarantee that a woman is not matched to more than one man, we use the following constraint:
$$
\ba l
  \lar  $\{$ \ii{marry}(x,y) : \ii{man}(x) $\}$ > 1, \ii{woman}(y).
\ea
$$
Individuals who stay single under the represented matching by \ii{marry} atoms are described by the \ii{msingle} and \ii{wsingle} atoms.
$$ 
\ba l
  \ii {msingle}(x) \lar \ii{man}(x), $\{$ \ii{marry}(x, y) : \ii {woman}(y) $\}0$.\\
\ii{wsingle}(y) \lar \ii{woman}(y), $\{$\ii {marry}(x, y) : \ii {man}(x) $\}0.$
\ea
$$

To establish stability, we refer to conditions A1--A3. The following set of constraints respectively describe and eliminate blocking pairs that are described by (a) -- (d) of A3. Conditions A1 and A2 also hold for each constraint.
$$
\ba l
 \lar \ii {acceptable}(x,y), \ii{msingle}(x), \ii{wsingle}(y). \\
\lar  \ii{wsingle}(y), \ii{marry}(x, y1),  \ii{mprefer}(x, y, y1), \ii{acceptable}(x, y). \\
\lar \ii{msingle}(x),  \ii{marry}(x1,y),  \ii{wprefer}(y, x, x1), \ii{acceptable}(x, y). \\
\lar  \ii{marry}(x,y1), \ii{marry}(x1,y),  \ii {mprefer}(x, y, y1), \ii {wprefer}(y, x, x1). \\
\ea
$$

Given the ASP formulation $P$ whose rules are described above and the ASP description $F_{I}$ of an \smti instance $I$, the ASP solver \clingo generates a stable matching.

We can transform the ASP program $P \cup F_I$ into an equivalent propositional theory $\Gamma(I)$ so that the answer sets for $P \cup F_I$ and the models for $\Gamma(I)$ are in a 1-1 correspondence~\cite{ErdemL03,LinZ04}. The ASP solver \cmodels~\cite{cmodels} is developed on this idea, and thus allows us to utilize SAT solvers (e.g., \zchaff~\cite{chaff}) to compute solutions for \smti instances.  An alternative SAT encoding of \smti can be obtained from the formulation of \citeN{Drummond15sat}.

To solve Sex-Equal \smti, we add the following weak constraint to optimize sex equality:
$$
\ba l
\xleftarrow{\backsim} t = \#\ii{sum}\{r1 - r2, x, y : \ii{marry}(x, y), \ii{mrank}(x, y, r1), \ii{wrank}(y, x, r2)\}.\ [|t|@1]
\ea
$$

To solve Egalitarian \smti, we add the following weak constraint to minimize the cost function:

$$
\ba l
\xleftarrow{\backsim}  \ii{marry}(x,y), \ii{mrank}(x,y,r1),\ii{wrank}(y,x,r2).\  [r1+r2@1,x,y]
\ea
$$

To solve Max Cardinality \smti, we add the following weak constraints to minimize the number of singles:
$$ 
\ba l
\xleftarrow{\backsim} \ii{wsingle}(x).\ [1@1,W,x] \\
\xleftarrow{\backsim} \ii{msingle}(y).\ [1@1,M,y] \\
\ea
$$

Note that combinations of these optimizations are possible, by giving priorities to weak constraints: instead of specifying just the weight $t$ of the weak constraints, we need to specify also their priorities $p$ by an expression $w@p$.  The weak constraints are supported by the ASP solver \clingo but not by \cmodels.


\section{Solving \smti using ILP and CP}
\label{sec:ilp}

We consider the ILP formulation of Max Cardinality \smti by Delorme et al.~\citeyear{DelormeGGKMP19} as a basis and introduce ILP formulations for Sex-Equal \smti and Egalitarian \smti.

Let us introduce the necessary notation to describe the ILP formulation.
\begin{itemize}
    \item $ML$: A list of lists of size $n$. $ML[i]$ represents a preference list of man $i$.
    \item $WL$: A list of lists of size $n$. $WL[j]$ represents a preference list of woman $j$.
    \item $ML_{{i}_{\leq}}(j)$: Set of men that woman $j$ ranks at the same level or better than man $i$.
    \item $WL_{j_\leq}(i)$: Set of women that man $i$ ranks at the same level or better than woman $j$.
    \item $mrank(i,j)$: The rank of woman $j$ in $ML[i]$.
    \item $wrank(j,i)$: The rank of man $i$ in $WL[j]$.
    \item $M$: A matrix of size $n \times n$ which denotes the matching where rows represent men and columns represent women.
   \item  $x_{ij}$: A binary variable that refers to an index of $M$, and is equal to 1 if man $i$ and woman $j$ are matched, 0 otherwise.
\end{itemize}

The acceptability constraint forces the variable $x_{ij}$ to be 0 if man $i$ and woman $j$ are not mutually acceptable:
$$ 
\ba l
$$ \{x_{ij} = 0: i \notin WL[j], j \notin ML[i]\}.$$
\ea
$$

The following constraint ensures that each man is matched to at most one woman. Since at most one $x_{ij}$ value can be equal to 1, sum of the cells in a row $i$ of $M$ should be at most one:
$$ 
\ba l
$$ \sum_{j \in WL} x_{ij} \leq 1.$$
\ea
$$
Similarly, the following constraint ensures each woman is matched to at most one man:
$$ 
\ba l
$$\sum_{i \in ML} x_{ij} \leq 1.$$
\ea
$$

The stability constraint ensures that there are no blocking pairs. The variable $q$ in the following constraint represents a member of the set of women who has the same or better rank than woman $j$ for a man $i$, and $p$ represents a member of the set of men who has the same or better rank than man $i$ for woman $j$. Hence, if man $i$ is matched to someone who is not $q$, or woman $j$ is matched to someone who is not $p$, man $i$ and woman $j$ form a blocking pair:
$$ 
\ba l
$$ 1-\sum_{q \in WL_{{j}_{\leq}}(i)}x_{iq} \leq \sum_{p \in ML_{{i}_{\leq}}(j)}x_{pj}.$$
\ea
$$

The following objective function represents the Max Cardinality optimization for the ILP model:
$$
\max (\sum_{i \in ML}\sum_{j \in WL} x_{ij}).
$$
\noindent Since the matrix $M$ denotes the matching and every matrix cell, denoted by variable $x_{ij}$, is 1 when the pair represented with that cell is matched; maximizing the sum of all cells corresponds to maximizing the number of matched pairs.

We use the following objective function for Sex-Equal optimization:
$$
\min(| \sum_{i \in ML} \sum_{j \in WL} (mrank(i,j) \times x_{ij} - wrank(j,i) \times x_{ij})|) .
$$
\noindent This objective function aims to minimize the difference between the sum of the ranks of matched pairs from men's perspective and the sum of the ranks of matched pairs from women's perspective.

The following objective function provides Egalitarian optimization:
$$
\min(\sum_{i \in ML} \sum_{j \in WL} (mrank(i,j) \times x_{ij} + wrank(j,i) \times x_{ij})) .
$$
\noindent In Egalitarian optimization, the total satisfaction of individuals is maximized. Since better preference yields lower rank, by minimizing sum of ranks of all matched pairs, the total satisfaction is maximized.

We implement the ILP models using Gurobi and Google OR-Tools MIP. With slight variations of the Google-OR Tools MIP implementation, we can also utilize Google-OR Tools CP.

\section{Solving Max Cardinality \smti using Local Search Algorithms}
\label{sec:local}

To compare with the ASP, ILP and CP approaches to solve Max Cardinality \smti, we also consider the existing local search algorithms.

\paragraph{\bf Random-Restart Stochastic Hill-Climbing Search.}

We consider the local search algorithm LTIU introduced by Gelain et al.~\citeyear{gelain2013local} to solve Max Cardinality \smti as the basis.  The input of this algorithm consists of the sets of $n$ men and $n$ women, preference lists of each man and woman, and a step or a time limit $T$.

LTIU performs a random-restart stochastic hill-climbing search (Algorithm~\ref{alg:ltiu}). The local search starts from a random matching.  The neighbors of a matching $\mu$ are all matchings obtained from $\mu$ by removing one ``undominated'' blocking pair.\footnote{Given two blocking pairs $(m,w)$ and $(m, w')$, $(m,w)$ dominates (from the men’s point of view) $(m,w')$ if $m$ prefers $w$ to~$w'$. It is also defined for the women’s point of view. A men- (resp., women-) undominated blocking pair is a blocking pair such that there is no other blocking pair that dominates it from the men’s (resp., women’s) point of view.} The objective function is the sum of the number of blocking pairs and the number of singles.

LTIU tries to minimize the value of this objective function: at each search step, it chooses a random matching that has the minimum evaluation value. If there exists no matching such that its evaluation value is less than the current matching, then the algorithm chooses a random matching from the neighborhood. Furthermore, with a certain probability, LTIU can choose a random matching in the neighborhood.

The algorithm terminates when it finds a perfect matching (i.e., a matching with no singles and no blocking pairs) or it reaches the time limit. If the algorithm does not find any stable matching during the search, then the best matching with a minimum evaluation value is returned.

During the search, a random restart is applied in order to avoid search near local minima (i.e., it reaches a matching with no blocking pair and thus, has an empty neighborhood).

\paragraph{\bf Genetic Algorithm.}

We consider the genetic algorithm (GA) introduced by \citeN{haas2020} to solve Max Cardinality \smti (Algorithm~\ref{alg:ga}). Unlike other methods, the algorithm generates stable pairs at every step to obtain matchings with improved properties.

According to the GA approach, chromosomes are solutions to the given \smti problem. In other words, a chromosome is a stable matching $\langle X,Y \rangle$ containing several genes which are pairs of matched women and men $\langle x_i ,y_i \rangle$. To create stable chromosomes for the initial population, ties are arbitrarily broken, then the Gale-Shapley algorithm, also known as deferred acceptance algorithm (DA), is used to find stable matchings.

The fitness function is constructed based on the optimization variant. For Max Cardinality \smti, a chromosome/matching $\mu$ is said to be fitter if the property $NumPairs = \sum_{\langle x,y \rangle} \{ \langle x,y \rangle | x \neq \emptyset \land y \neq \emptyset \}$ is higher for that matching.

To improve the fitness of the chromosome/matching to the desired outcome, two genetic operators are used: the cycle crossover operator and the mutation operator. The fittest individuals are chosen randomly at each iteration to mate and bear offspring. Initially, individuals are randomly selected based on their ranks. The formula for calculating the chance to select an individual (for population $P$ and individual $i$) is as follows:
$$
chance(i) = \frac{fitness(i)}{\sum_{k \in P} fitness(k)}
.
$$
By integrating two parents’ solutions, the cycle crossover operator produces new possible solutions by selecting a sequence of two or more pairs and flips either $x$ or $y$ for each pair in the sequence.

The mutation operator searches for Pareto-improvement cycles as described by \citeN{erdil2008s} to increase the number of matched participants, provided a certain mutation probability.

Our implementation of GA concentrates solely on stable allocations for the Two-Sided Matching application. Particularly, as part of crossover and mutation operators, we have included additional controls that define possible blocking pairs introduced by the changes and only consider adjustments to be acceptable if they do not lead to blocking pairs to ensure that the genetic algorithm returns valid and stable allocations \cite{haas2020}.

\section{Experiments}\label{sec:experiments}

We have empirically compared different methods to solve \smti and its hard variants.

\paragraph{\bf Experimental Setup.}

To test and analyze the models and our implementations, we need an instance generator. For this, we have implemented the random instance generator proposed by \citeN{gent2002empirical}.

The random instance generator takes 3 inputs to generate instances: instance size $n$, (i.e, number of men and women), probability of incompleteness $p1$, and probability of ties $p2$. We have generated a benchmark set, with instance sizes $n=50$ and $n=100$, where the value of $p1$ changes in the range of [0.1, 0.8] and the value of $p2$ changes in the range [0.1, 0.9] with 0.1 step. There are 144 combinations and for each combination, we have generated 10 instances and averaged the results.

The tests were run on a local machine which has Ubuntu 18.04.1 as operating system and x86\_64 processor. The algorithms are implemented in Python programming language with version 3.6.9. Additionally, we use Gurobi version 9.1.1, OR-Tools version 8.1.8487, \clingo version 5.2.2, \cmodels version 3.79 with the SAT solver \zchaff 2007.3.12, and SAT-E version released on May 17, 2016 with the SAT solver \lingeling~bcj.

We set the time limit for each solver to 2000 seconds, memory limit to 2 GB, and step limits for LTIU and GA implementations to 50000, and 10000, respectively. In GA experiments, the population size is set to 50, the number of evolution rounds is set to 1000, and the mutation probability is set to $0.2$. In LTIU experiments, we set the probability of selecting a random matching from the neighborhood to $0.2$.

\begin{table}[t]
    \caption{\smti: Average CPU-Times (in seconds) for varying $p1$ and $p2$ values with $n=50$}
     \label{tab:smti50}
    \begin{minipage}{\textwidth}
     {\footnotesize    \centering
     \begin{tabular}{ccccccccccc}
    \hline\hline
      &  &  \multicolumn{9}{c}{$p2$} \\
      \cline{3-11}\\
       {Solver} & $p1$ & 0.1 & 0.2 & 0.3 & 0.4 & 0.5 & 0.6 & 0.7 & 0.8 & 0.9 \\
      \hline
      \multirow{8}{*}{\clingo} & 0.1 & 5.07 & 4.99 & 4.96 & 4.82 & 4.74 & 4.61 & 4.36 & 4.03 & 3.09\\
      & 0.2 & 3.41 & 3.39 & 3.33 & 3.27 & 3.17 & 3.04 & 2.91 & 2.63 & 1.88\\
     & 0.3 & 2.11 & 2.05 & 2.02 & 2.01 & 1.92 & 1.86 & 1.73 & 1.51 & 1.14\\
      & 0.4 & 1.23 & 1.2 & 1.17 & 1.19 & 1.12 & 1.09 & 1.0 & 0.92 & 0.63\\
     & 0.5 & 0.68 & 0.65 & 0.65 & 0.64 & 0.61 & 0.6 & 0.51 & 0.45 & 0.34\\
    & 0.6 & 0.35 & 0.33 & 0.35 & 0.32 & 0.3 & 0.29 & 0.26 & 0.22 & 0.17\\
     & 0.7 & 0.16 & 0.15 & 0.15 & 0.14 & 0.14 & 0.14 & 0.12 & 0.11 & 0.09\\
    & 0.8 & 0.06 & 0.07 & 0.06 & 0.06 & 0.07 & 0.06 & 0.05 & 0.05 & 0.03\\
       \hline
      \multirow{8}{*}{\cmodels}
   & 0.1 & 7.29 & 7.18 & 7.18 & 6.93 & 6.62 & 6.44 & 5.97 & 5.41 & 4.38\\
   & 0.2 & 4.92 & 4.95 & 4.85 & 4.7 & 4.6 & 4.2 & 4.04 & 3.7 & 2.95\\
   & 0.3 & 3.34 & 3.31 & 3.23 & 3.17 & 3.03 & 3.01 & 2.68 & 2.41 & 1.88\\
   & 0.4 & 2.02 & 1.99 & 1.97 & 1.98 & 1.89 & 1.83 & 1.64 & 1.53 & 1.11\\
   & 0.5 & 1.19 & 1.12 & 1.16 & 1.13 & 1.06 & 1.07 & 0.89 & 0.81 & 0.61\\
   & 0.6 & 0.61 & 0.57 & 0.6 & 0.56 & 0.51 & 0.5 & 0.46 & 0.39 & 0.28\\
   & 0.7 & 0.26 & 0.25 & 0.24 & 0.23 & 0.23 & 0.22 & 0.19 & 0.17 & 0.13\\
   & 0.8 & 0.1 & 0.1 & 0.1 & 0.09 & 0.1 & 0.09 & 0.08 & 0.08 & 0.07\\
    \hline
      \multirow{8}{*}{SAT-E}
      & 0.1 & 0.79 & 0.84 & 0.86 & 0.88 & 0.93 & 0.94 & 0.98 & 0.98 & 1.0\\
    & 0.2 & 0.62 & 0.67 & 0.68 & 0.7 & 0.74 & 0.78 & 0.78 & 0.84 & 0.82\\
    & 0.3 & 0.51 & 0.54 & 0.55 & 0.58 & 0.6 & 0.65 & 0.66 & 0.68 & 0.69\\
    & 0.4 & 0.4 & 0.41 & 0.42 & 0.44 & 0.46 & 0.48 & 0.5 & 0.53 & 0.56\\
    & 0.5 & 0.31 & 0.32 & 0.33 & 0.34 & 0.35 & 0.36 & 0.36 & 0.39 & 0.4\\
    & 0.6 & 0.24 & 0.25 & 0.26 & 0.26 & 0.27 & 0.28 & 0.28 & 0.28 & 0.29\\
    & 0.7 & 0.18 & 0.18 & 0.18 & 0.19 & 0.19 & 0.2 & 0.2 & 0.2 & 0.2\\
    & 0.8 & 0.13 & 0.13 & 0.13 & 0.14 & 0.14 & 0.15 & 0.14 & 0.15 & 0.15\\

       \hline \hline
    \end{tabular}}
    \end{minipage}
    \vspace{-\baselineskip}
\end{table}

\begin{table}[t]
    \caption{\smti: Average CPU-Times (in seconds) for varying $p1$ and $p2$ values with $n=100$}
     \label{tab:smti100}
   \begin{minipage}{\textwidth}
    {\footnotesize    \centering
    \begin{tabular}{ccccccccccc}
    \hline\hline
      &  &  \multicolumn{9}{c}{$p2$} \\
            \cline{3-11}\\
       {Solver} & $p1$ & 0.1 & 0.2 & 0.3 & 0.4 & 0.5 & 0.6 & 0.7 & 0.8 & 0.9 \\
      \hline
      \multirow{8}{*}{\clingo}  & 0.1 & 120.5 & 116.94 & 122.21 & 121.05 & 107.89 & 108.29 & 104.07 & 95.7 & 81.07 \\
      & 0.2 & 80.14 & 80.63 & 78.42 & 79.3 & 74.38 & 69.78 & 68.89 & 66.07 & 54.74\\
      & 0.3 & 45.37 & 45.36 & 43.66 & 43.22 & 40.64 & 38.17 & 39.63 & 34.1 & 29.4\\
     & 0.4 & 27.52 & 26.65 & 26.15 & 25.84 & 25.27 & 23.86 & 22.75 & 21.76 & 17.96\\
     & 0.5 & 14.52 & 14.06 & 14.13 & 13.99 & 13.93 & 13.13 & 13.29 & 11.21 & 9.64\\
     & 0.6 & 7.12 & 7.16 & 7.2 & 7.02 & 6.87 & 7.53 & 6.17 & 5.83 & 4.63\\
     & 0.7 & 3.12 & 3.19 & 3.12 & 3.12 & 3.08 & 3.09 & 2.8 & 2.41 & 1.87\\
     & 0.8 & 0.95 & 0.96 & 0.97 & 0.92 & 0.93 & 0.84 & 0.74 & 0.64 & 0.46\\

       \hline
      \multirow{8}{*}{\cmodels}
      & 0.1 & M & M & M & M & M & M & M & M & M\\
      & 0.2 & M & M & M & M & M & M & M & M & M\\
      & 0.3 & M & M & M & M & M & M & M & M & M\\
      & 0.4 & M & M & M & M & M & M & M & M & M\\
      & 0.5 & 23.29 & 23.33 & 23.49 & 23.66 & 23.91 & 23.77 & 21.38 & 17.03 & 14.37\\
      & 0.6 & 12.63 & 12.93 & 12.97 & 12.62 & 11.95 & 11.72 & 10.22 & 8.64 & 6.71\\
      & 0.7 & 4.62 & 4.85 & 4.66 & 4.73 & 4.87 & 4.62 & 3.88 & 3.44 & 2.71\\
      & 0.8 & 1.47 & 1.42 & 1.45 & 1.4 & 1.4 & 1.28 & 1.16 & 1.01 & 0.76\\

      \hline
      \multirow{8}{*}{SAT-E}
       & 0.1 & 6.04 & 5.78 & 6.79 & 6.84 & 6.96 & 7.48 & 8.0 & 7.24 & 5.99\\
      & 0.2 & 4.91 & 4.92 & 5.02 & 5.98 & 5.73 & 5.82 & 6.57 & 7.68 & 4.69\\
      & 0.3 & 3.87 & 3.98 & 4.67 & 4.54 & 5.09 & 4.09 & 4.02 & 4.83 & 4.07\\
      & 0.4 & 2.74 & 3.21 & 3.41 & 3.37 & 4.14 & 3.56 & 3.93 & 3.41 & 3.16\\
      & 0.5 & 1.96 & 1.97 & 2.1 & 2.28 & 2.22 & 2.26 & 2.54 & 2.21 & 2.67\\
      & 0.6 & 1.24 & 1.33 & 1.4 & 1.38 & 1.4 & 1.5 & 1.66 & 1.8 & 1.46\\
      & 0.7 & 0.74 & 0.8 & 0.79 & 0.83 & 0.89 & 0.92 & 0.99 & 0.96 & 0.96\\
      & 0.8 & 0.41 & 0.42 & 0.44 & 0.45 & 0.46 & 0.47 & 0.49 & 0.51 & 0.53\\

       \hline \hline
    \end{tabular}}

    \footnotesize{M: Memory limit reached (over 2 GB)}
    \end{minipage}
    \vspace{-\baselineskip}
\end{table}

\begin{table}[t]
    \caption{Max Cardinality \smti: Average CPU-Times (in seconds) for $n=50$.}
    \label{tab:maxCard50}
    \begin{minipage}{\textwidth}
    {\footnotesize    \centering
    \begin{tabular}{ccccccccccc}
    \hline\hline
              &  &  \multicolumn{9}{c}{$p2$} \\
                    \cline{3-11}\\
      Solver &$p1$ & 0.1 & 0.2 & 0.3 & 0.4 & 0.5 & 0.6 & 0.7 & 0.8 & 0.9 \\
      \hline
      \multirow{8}{*}{\clingo} & 0.1 & 5.23 & 5.28 & 5.1 & 5.06 & 4.98 & 4.81 & 4.54 & 4.23 & 3.29 \\
      & 0.2 & 3.57 & 3.6 & 3.54 & 3.41 & 3.4 & 3.23 & 3.11 & 2.83 & 2.08 \\
      & 0.3 & 2.3 & 2.23 & 2.18 & 2.21 & 2.09 & 2 & 0.28 & 1.67 & 1.3 \\
      & 0.4 & 1.36 & 1.32 & 1.27 & 1.35 & 1.24 & 1.22 & 0.26 & 1.05 & 0.75 \\
      & 0.5 & 0.77 & 0.74 & 0.76 & 0.74 & 0.71 & 0.7 & 0.5 & 0.55 & 0.41 \\
      & 0.6 & 0.41 & 0.39 & 0.41 & 0.39 & 0.35 & 0.35 & 0.2 & 0.29 & 0.21 \\
      & 0.7 & 0.19 & 0.19 & 0.19 & 0.18 & 0.18 & 0.18 & 5.64 & 0.14 & 0.11 \\
      & 0.8 & 0.09 & 0.09 & 0.09 & 0.09 & 0.09 & 0.08$^{*}$ & 65.11 & 0.07 & 0.06 \\
       \hline
      \multirow{8}{*}{Gurobi} & 0.1 & 0.44 & 0.47 & 0.48 & 0.5 & 0.49 & 0.52 & 0.58 & 0.66 & 0.59 \\
      & 0.2 & 0.34 & 0.35 & 0.37 & 0.39 & 0.4 & 0.44 & 0.46 & 0.54 & 0.51 \\
      & 0.3 & 0.27 & 0.29 & 0.32 & 0.37 & 0.33 & 0.36 & 0.38 & 0.41 & 0.39 \\
      & 0.4 & 0.21 & 0.22 & 0.22 & 0.24 & 0.26 & 0.28 & 0.29 & 0.33 & 0.32 \\
      & 0.5 & 0.17 & 0.16 & 0.17 & 0.18 & 0.19 & 0.21 & 0.21 & 0.23 & 0.24 \\
      & 0.6 & 0.12 & 0.12 & 0.14 & 0.14 & 0.14 & 0.15 & 0.16 & 0.17 & 0.16 \\
      & 0.7 & 0.09 & 0.09 & 0.1 & 0.1 & 0.11 & 0.11 & 0.12 & 0.12 & 0.11 \\
      & 0.8 & 0.07 & 0.07 & 0.07 & 0.08 & 0.08 & 0.08 & 0.08 & 0.09 & 0.09 \\
       \hline
      \multirow{8}{*}{OR-MIP} & 0.1 & 0.69 & 0.71 & 0.72 & 0.72 & 0.73 & 0.74 & 0.77 & 0.8 & 0.84 \\
      & 0.2 & 0.55 & 0.57 & 0.58 & 0.6 & 0.6 & 0.61 & 0.62 & 0.65 & 0.7 \\
      & 0.3 & 0.44 & 0.46 & 0.48 & 0.48 & 0.49 & 0.5 & 0.52 & 0.53 & 0.56 \\
      & 0.4 & 0.35 & 0.36 & 0.36 & 0.38 & 0.39 & 0.39 & 0.4 & 0.43 & 0.45 \\
      & 0.5 & 0.27 & 0.27 & 0.29 & 0.29 & 0.3 & 0.31 & 0.31 & 0.33 & 0.36 \\
      & 0.6 & 0.21 & 0.2 & 0.22 & 0.23 & 0.23 & 0.24 & 0.24 & 0.25 & 0.27 \\
      & 0.7 & 0.15 & 0.15 & 0.16 & 0.16 & 0.17 & 0.18 & 0.18 & 0.19 & 0.19 \\
      & 0.8 & 0.11 & 0.11 & 0.11 & 0.12 & 0.12 & 0.13 & 0.13 & 0.14 & 0.14 \\
        \hline
      \multirow{8}{*}{OR-CP} & 0.1 & 0.9 & 0.93 & 0.91 & 1.03 & 1.05 & 1.07 & 1.11 & 1.14 & 1.32 \\
      & 0.2 & 0.74 & 0.73 & 0.75 & 0.77 & 0.81 & 0.85 & 1.03 & 1.0 & 0.99 \\
      & 0.3 & 0.59 & 0.6 & 0.62 & 0.64 & 0.65 & 0.73 & 0.81 & 0.75 & 0.76 \\
      & 0.4 & 0.47 & 0.47 & 0.48 & 0.5 & 0.51 & 0.6 & 1.61 & 0.75 & 0.6 \\
      & 0.5 & 0.38 & 0.37 & 0.38 & 0.41 & 0.44 & 0.46 & 0.53 & 0.54 & 0.53 \\
      & 0.6 & 0.3 & 0.29 & 0.31 & 0.33 & 0.34 & 0.41 & 0.52 & 0.81 & 0.4 \\
      & 0.7 & 0.24 & 0.23 & 0.24 & 0.26 & 0.3 & 0.4 & 0.48 & 0.58 & 0.31 \\
      & 0.8 & 0.18 & 0.18 & 0.19 & 0.19 & 0.21 & 0.88 & 0.29 & 0.26 & 0.39 \\
       \hline     \hline
    \end{tabular}}

\footnotesize{* 1 instance reached time limit (over 2000 seconds)}
 \vspace{-\baselineskip}
    \end{minipage}
\end{table}

\begin{table}[t]
    \caption{Max Cardinality \smti: Average CPU-Times (in seconds) for varying $p1$ and $p2$ values with $n=50$.}
    \label{tab:maxCardLocalCPU50}
    \begin{minipage}{\textwidth}
    {\footnotesize    \centering
    \begin{tabular}{ccccccccccc}
    \hline\hline
      &  &  \multicolumn{9}{c}{$p2$} \\
            \cline{3-11}\\
      Solver & $p1$ & 0.1 & 0.2 & 0.3 & 0.4 & 0.5 & 0.6 & 0.7 & 0.8 & 0.9 \\
       \hline
       \multirow{8}{*}{LTIU} & 0.1 & 0.46 & 0.51 & 0.4 & 0.47 & 0.4 & 0.39 & 0.36 & 0.33 & 0.24 \\
      & 0.2 & 0.36 & 0.4 & 0.36 & 0.41 & 0.36 & 0.33 & 0.36 & 0.28 & 0.22 \\
      & 0.3 & 0.36 & 0.39 & 0.37 & 0.33 & 0.3 & 0.31 & 0.28 & 0.26 & 0.21 \\
      & 0.4 & 43.33 & 0.33 & 0.28 & 0.9 & 0.29 & 0.25 & 0.26 & 0.22 & 0.18 \\
      & 0.5 & 15.57 & 0.27 & 2.45 & 0.27 & 0.34 & 0.24 & 0.5 & 0.38 & 0.15 \\
      & 0.6 & 16.71 & 29.4 & 6.27 & 12.75 & 0.54 & 0.27 & 0.2 & 0.19 & 0.31 \\
      & 0.7 & 47.9 & 43.58 & 82.46 & 14.97 & 29.92 & 397 & 5.64 & 14 & 1.08 \\
      & 0.8 & 114.67 & 94.89 & 85.28 & 93.76 & 77.39 & 54.16 & 65.11 & 37.17 & 17.27 \\
      \hline
      \multirow{8}{*}{GA} & 0.1 & 0.05 & 0.07 & 0.09 & 0.1 & 0.11 & 0.11 & 0.11 & 0.11 & 0.1  \\
      & 0.2 & 0.05 & 0.07 & 0.08 & 0.09 & 0.1 & 0.1 & 0.1 & 0.1 & 0.09 \\
      & 0.3 & 0.04 & 0.06 & 0.07 & 0.08 & 0.09 & 0.09 & 0.09 & 0.09 & 0.8 \\
      & 0.4 & 20.98 & 0.05 & 0.06 & 0.07 & 0.08 & 0.08 & 0.08 & 0.08 & 0.07 \\
      & 0.5 & 11.25 & 0.04 & 0.05 & 0.06 & 0.06 & 0.07 & 0.07 & 0.06 & 0.06 \\
      & 0.6 & 9.89 & 19.09 & 0.04 & 9.94 & 0.05 & 0.05 & 0.05 & 0.05 & 0.05 \\
      & 0.7 & 25.57 & 17.72 & 43.6 & 0.04 & 8.65 & 0.04 & 0.04 & 0.04 & 0.04 \\
      & 0.8 & 67.74 & 61.75 & 55.28 & 40.64$^{\dag}$ & 26.61$^{\dag}$ & 19.83 & 12.52$^{\dag}$ & 12.01 & 0.03$^{\dag}$ \\
      \hline \hline
    \end{tabular}}

    \footnotesize{\dag \hspace{0.25mm} 1 instance stopped execution}
\end{minipage}
 \vspace{-\baselineskip}
\end{table}

\begin{table}[t]
     \caption{Max Cardinality \smti: Average CPU-Times (in seconds) for $n=100$.}
    \label{tab:maxCard100}
    \begin{minipage}{\textwidth}
    {\footnotesize    \centering
    \begin{tabular}{ccccccccccc}
    \hline\hline
      &  &  \multicolumn{9}{c}{$p2$} \\
            \cline{3-11}\\
      Solver & $p1$ & 0.1 & 0.2 & 0.3 & 0.4 & 0.5 & 0.6 & 0.7 & 0.8 & 0.9 \\
      \hline
      \multirow{8}{*}{\clingo} & 0.1 & 105.7 & 106.9 & 101.52 & 97.56 & 100.57 & 95.27 & 92.78 & 88.63 & 75.95 \\
      & 0.2 & 72.22 & 71.3 & 72.29 & 70.94 & 68.72 & 65.52 & 62.66 & 60.11 & 50.64 \\
      & 0.3 & 41.71 & 41.86 & 72.29 & 40.62 & 39 & 37.3 & 37 & 32.96 & 27.92 \\
      & 0.4 & 25.76 & 25.78 & 25.29 & 24.93 & 24.34 & 23.68 & 21.62 & 20.63 & 17.59 \\
      & 0.5 & 14.34 & 14.45 & 14.2 & 14.09 & 14.1 & 12.99 & 12.73 & 11.2 & 9.55 \\
      & 0.6 & 7.24 & 7.34 & 7.3 & 7.02 & 6.95 & 7.18 & 6.25 & 5.73 & 4.55 \\
      & 0.7 & 2.98 & 3.07 & 2.97 & 2.92 & 3.02 & 2.87 & 2.68 & 2.29 & 1.81 \\
      & 0.8 & 0.92 & 0.89 & 0.9 & 0.86 & 0.87 & 0.81 & 0.75 & 0.64 & 0.46 \\
       \hline
      \multirow{8}{*}{Gurobi} & 0.1 & 4.66 & 4.76 & 4.84 & 4.99 & 5.17 & 5.37 & 5.74 & 6.01 & 6.17 \\
       & 0.2 & 3.51 & 3.58 & 3.66 & 3.91 & 3.97 & 4.25 & 4.46 & 4.38 & 4.53 \\
       & 0.3 & 2.53 & 2.6 & 2.7 & 2.86 & 3.0 & 3.07 & 3.22 & 3.35 & 3.48 \\
       & 0.4 & 1.79 & 1.84 & 1.92 & 1.95 & 2.12 & 2.19 & 2.23 & 2.36 & 2.4 \\
       & 0.5 & 1.25 & 1.25 & 1.34 & 1.37 & 1.49 & 1.58 & 1.66 & 1.65 & 1.71 \\
       & 0.6 & 0.81 & 0.83 & 0.88 & 0.95 & 1.02 & 1.08 & 1.21 & 1.2 & 1.19 \\
       & 0.7 & 0.53 & 0.55 & 0.56 & 0.6 & 0.66 & 0.71 & 0.77 & 0.79 & 0.76 \\
       & 0.8 & 0.35 & 0.35 & 0.38 & 0.39 & 0.41 & 0.45 & 0.46 & 0.48 & 0.45 \\
       \hline
      \multirow{8}{*}{OR-MIP} & 0.1 & 5.91 & 5.98 & 6.02 & 6.17 & 6.2 & 6.22 & 6.31 & 6.56 & 6.75\\
       & 0.2 & 4.68 & 4.75 & 4.77 & 4.88 & 4.87 & 4.88 & 4.98 & 5.09 & 5.37 \\
       & 0.3 & 3.59 & 3.65 & 3.73 & 3.77 & 3.84 & 3.77 & 3.86 & 3.94 & 4.05\\
       & 0.4 & 2.64 & 2.7 & 2.72 & 2.76 & 2.81 & 2.74 & 2.74 & 2.85 & 3.0 \\
       & 0.5 & 1.86 & 1.88 & 1.92 & 1.94 & 1.99 & 1.98 & 2.01 & 2.03 & 2.18 \\
       & 0.6 & 1.27 & 1.31 & 1.35 & 1.39 & 1.38 & 1.4 & 1.41 & 1.45 & 1.53 \\
       & 0.7 & 0.83 & 0.86 & 0.88 & 0.92 & 0.94 & 0.94 & 0.95 & 0.94 & 1.0 \\
       & 0.8 & 0.54 & 0.55 & 0.58 & 0.59 & 0.62 & 0.62 & 0.61 & 0.62 & 0.64 \\
        \hline
      \multirow{8}{*}{OR-CP} & 0.1 & 7.63 & 7.98 & 8.16 & 8.41 & 8.81 & 9.77 & 13.72 & 13.91 & 26.17 \\
       & 0.2 & 6.57 & 6.67 & 6.83 & 7.59 & 7.47 & 7.13 & 8.89 & 12.4 & 18.25 \\
        & 0.3 & 4.66 & 4.81 & 5.65 & 5.81 & 6.21 & 12.48 & 9.86 & 19.66 & 20.61 \\
        & 0.4 & 3.54 & 3.73 & 4.01 & 4.27 & 6.88 & 6.1 & 31.63 & 25.45 & 13.72 \\
        & 0.5 & 2.6 & 2.67 & 3.16 & 3.29 & 5.16 & 8.3 & 6.18 & 4.99 & 10.15 \\
        & 0.6 & 1.86 & 1.92 & 2.1 & 3.06 & 5.61 & 7.67 & 9.82 & 6.13 & 6.37 \\
        & 0.7 & 1.27 & 1.33 & 1.55 & 3.48 & 4.65 & 13.28 & 6.64 & 6.88 & 3.37 \\
        & 0.8 & 0.89 & 0.9 & 1.11 & 1.59 & 17.22 & 72.66 & 195.72 & 15.32 & 1.6 \\
       \hline \hline
    \end{tabular}}
   \end{minipage}
    \vspace{-\baselineskip}
\end{table}

\begin{table}[t]
    \caption{Max Cardinality \smti: Average CPU-Times (in seconds) for varying $p1$ and $p2$ values with $n=100$.}
    \label{tab:maxCardLocalCPU100}
       \vspace{-.3\baselineskip}
    \begin{minipage}{\textwidth}
    {\footnotesize    \centering
    \begin{tabular}{ccccccccccc}
     \hline \hline
      & &  \multicolumn{9}{c}{$p2$} \\
            \cline{3-11}\\
      Solver & $p1$ & 0.1 & 0.2 & 0.3 & 0.4 & 0.5 & 0.6 & 0.7 & 0.8 & 0.9 \\
       \hline
      \multirow{8}{*}{LTIU} & 0.1 & 5.69 & 5.46 & 5.83 & 5.29 & 5.65 & 4.84 & 5.01 & 4.25 & 3.69 \\
      & 0.2 & 4.90 & 4.94 & 5.2 & 4.96 & 5.09 & 4.6 & 4.55 & 3.8 & 3.26 \\
      & 0.3 & 4.45 & 4.55 & 4.36 & 4.42 & 4.47 & 4.17 & 4.04 & 3.48 & 2.93 \\
      & 0.4 & 3.86 & 3.83 & 3.93 & 3.69 & 3.78 & 3.78 & 3.3 & 3.02 & 2.9 \\
      & 0.5 & 3.5 & 95.63 & 3.42 & 3.3 & 3.29 & 2.97 & 2.88 & 2.96 & 2.37 \\
      & 0.6 & 2.94 & 83.01 & 6.05 & 2.6 & 12.54 & 2.66 & 2.56 & 2.15 & 1.72\\
      & 0.7 & 141.2 & 126.92 & 80.2 & 16.23 & 7.5 & 3.88 & 4.79 & 4.92 & 2.09 \\
      & 0.8 & 420.05 & 529.24 & 235.42 & 245.69 & 244.11 & 136.01 & 111.51 & 88.17 & 17.62 \\
      \hline
      \multirow{8}{*}{GA} & 0.1 & 0.21 & 0.29 & 0.36 & 0.4 & 0.44 & 0.45 & 0.44 & 0.41 & 0.37  \\
      & 0.2 & 0.19 & 0.26 & 0.32 & 0.36 & 0.39 & 0.4 & 0.39 & 0.37 & 0.33 \\
      & 0.3 & 0.17 & 0.23 & 0.28 & 0.32 & 0.34 & 0.35 & 0.35 & 0.33 & 0.29 \\
      & 0.4 & 0.15 & 0.2 & 0.25 & 0.27 & 0.3 & 0.31 & 0.3 & 0.28 & 0.26 \\
      & 0.5 & 0.12 & 0.17 & 0.2 & 0.23 & 0.25 & 0.25 & 0.25 & 0.24 & 0.022 \\
      & 0.6 & 0.1 & 0.14 & 0.17 & 0.18 & 0.2 & 0.21 & 0.2 & 0.19 & 0.18 \\
      & 0.7 & 54.38 & 0.11 & 0.13 & 0.14 & 0.15 & 0.16 & 0.16 & 0.15 & 0.14 \\
      & 0.8 & 306.54 & 261.9 & 87.45 & 86.7 & 43.75 & 38.43 & 0.11 & 0.11 & 0.1 \\
      \hline \hline
    \end{tabular}}
     \end{minipage}
      \vspace{-\baselineskip}
\end{table}

\paragraph{\bf Results and Discussion.}
For \smti, we have compared two approaches using three implementations: ASP (using \clingo), SAT (using \cmodels with \zchaff), and SAT (using SAT-E with \lingeling). For thand ese experiments, we have adapted the ASP formulation presented above, to the input language of \cmodels, and the SAT-E implementation to consider ties.

For each solver, for each combination of $p1$ and $p2$, the average CPU times are reported in Table~\ref{tab:smti50} for $n=50$, and in Table~\ref{tab:smti100} for $n=100$. We have observed that for all three implementations, the computation time decreases as $p1$ (i.e., probability of incompleteness) increases. As $p2$ (i.e., probability of ties) increases, the computation time decreases for \clingo and \cmodels, while it slightly increases for SAT-E.
\clingo and \cmodels are comparable for $p1\geq 0.5$ but \clingo is more advantageous than \cmodels due to less consumption of memory for $p1< 0.5$. In general, SAT-E is more advantageous than \cmodels, due to smaller theory sizes (Table~\ref{tab:smti50-atoms} and Table~ \ref{tab:smti50-clauses}) and a more efficient SAT solver. For instance,
for $n=50$ and $p1=p2=0.1$, SAT-E generates a propositional theory with
9022 atoms and 124392 clauses in average, whereas \cmodels generates a propositional theory with 113425 atoms and 1296263 clauses in average. According to a survey about the SAT solver competitions~\cite{JarvisaloBRS12}, \lingeling\ performs significantly better than \zchaff.

For Max Cardinality, we have compared four methods using six implementations: ASP (using \clingo), ILP (using Gurobi and Google-OR Tools MIP), CP (Google-OR Tools CP), and local search (LTIU and GA).

For each solver, for each combination of $p1$ and $p2$, the average CPU times are reported in Tables~\ref{tab:maxCard50} and~\ref{tab:maxCardLocalCPU50} for  $n=50$, and in Tables~\ref{tab:maxCard100} and~\ref{tab:maxCardLocalCPU100} for $n=100$. We have observed that, for a pair of $p1$ and $p2$ values, the average CPU time required for each solver to solve the instances increases with the value of $n$. This is due to the increase in the number of constraints to be satisfied in ILP approaches (Gurobi and OR-Tools solvers), larger space of matchings to search in the local search approaches (LTIU and GA), and the larger program size for ASP (\clingo).

For a pair of $n$ and $p2$ values, as $p1$ increases, the number of blocking pairs most likely decreases, and thus the average CPU time usually increases for local search methods (due to larger space of matchings). Meanwhile, the preference lists get shorter, and thus the average CPU time usually decreases for the other approaches (due to less number of constraints / rules).

For a pair of $n$ and $p1$ values, as $p2$ increases, the number of ties increases, and thus the average CPU time usually decreases for the local search methods (due to larger possibility of stable matchings, and, in addition, due to more variety in the initial population for GA). Meanwhile, the number of blocking pairs most likely increases, and thus the average CPU time usually increases for ILP and CP methods (due to larger constraints). The rules / constraints in ASP do not get larger, but the number of stability constraints increases, and thus the average CPU time usually decreases.

\begin{figure}
\vspace{0.5\baselineskip}
      \includegraphics[width=130mm,scale=0.65]{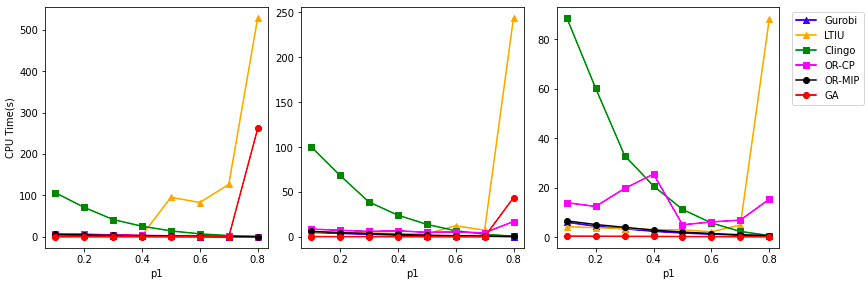}
     \vspace{-.5\baselineskip}
    \caption{CPU Time(s) for solving Max Cardinality \smti $p2=0.2$, $0.5$, and $0.8$ with $n=100$}
    \label{fig:maxcard}
     \vspace{-\baselineskip}
  \end{figure}

Figure~\ref{fig:maxcard} provides a comparison of the different approaches with respect to the CPU time for $n=100$. It can be observed that, for most instances, these approaches are comparable to each other, and that the local search methods takes more time for larger values of $p1$ (when the preference lists are less incomplete). We can also observe in that, for smaller values of $p1<0.7$, GA is more efficient, whereas, for larger values of $p1\geq 0.7$, Gurobi, Google-OR Tools MIP and \clingo are more efficient.

\begin{table}[t]
     \caption{Sex Equal \smti: Average CPU-Times (in seconds) for $n=50$.}
    \label{tab:optimizationS}
       \vspace{-.3\baselineskip}
    \begin{minipage}{\textwidth}
    {\footnotesize    \centering
    {\begin{tabular}{p{1ex}p{2ex}p{7ex}p{7ex}p{7ex}p{8ex}p{8ex}p{8ex}p{8ex}p{8ex}p{7ex}}
     \hline \hline
     \multirow{2}{*}{\rotatebox{90}{Solver}}&  &  \multicolumn{9}{c}{$p2$} \\
           \cline{3-11}\\
     & $p1$ & 0.1 & 0.2 & 0.3 & 0.4 & 0.5 & 0.6 & 0.7 & 0.8 & 0.9 \\
    \hline
     \multirow{8}{*}{\rotatebox{90}{Gurobi}}  & 0.1 & 0.48 & 0.56 & 0.55 & 0.6 & 0.64 & 0.71 & 0.81 & 0.8 & 0.77 \\
    & 0.2 & 0.39 & 0.44 & 0.48 & 0.63 & 0.55 & 0.74 & 0.76 & 0.93 & 0.71 \\
    & 0.3 & 0.31 & 0.35 & 0.38 & 0.46 & 0.53 & 0.52 & 0.64 & 0.6 & 0.58 \\
    & 0.4 & 0.25 & 0.27 & 0.28 & 0.32 & 0.37 & 0.38 & 0.44 & 0.47 & 0.46 \\
    & 0.5 & 0.2 & 0.21 & 0.25 & 0.26 & 0.3 & 0.37 & 0.29 & 0.35 & 0.35 \\
    & 0.6 & 0.16 & 0.16 & 0.18 & 0.21 & 0.2 & 0.23 & 0.25 & 0.24 & 0.23 \\
    & 0.7 & 0.12 & 0.13 & 0.14 & 0.15 & 0.15 & 0.17 & 0.17 & 0.18 & 0.17 \\
    & 0.8 & 0.1 & 0.1 & 0.1 & 0.11 & 0.12 & 0.12 & 0.12 & 0.13 & 0.13 \\
    \hline
    \multirow{8}{*}{\rotatebox{90}{OR-MIP}} & 0.1 & 0.84 & 0.84 & 0.86 & 0.87 & 0.93 & 0.98 & 1.08 & 1.36 & 1.95 \\
    & 0.2 & 0.67 & 0.69 & 0.71 & 0.74 & 0.74 & 0.87 & 0.91 & 1.51 & 2.36 \\
    & 0.3 & 0.54 & 0.55 & 0.56 & 0.58 & 0.63 & 0.67 & 0.7 & 1.01 & 1.21 \\
    & 0.4 & 0.45 & 0.44 & 0.44 & 0.48 & 0.51 & 0.52 & 0.64 & 0.94 & 1.35 \\
    & 0.5 & 0.35 & 0.36 & 0.37 & 0.38 & 0.38 & 0.51 & 0.51 & 0.75 & 1.06 \\
    & 0.6 & 0.27 & 0.27 & 0.29 & 0.32 & 0.32 & 0.41 & 0.52 & 0.46 & 0.47 \\
    & 0.7 & 0.21 & 0.21 & 0.22 & 0.23 & 0.24 & 0.26 & 0.31 & 0.42 & 0.34 \\
    & 0.8 & 0.16 & 0.17 & 0.18 & 0.18 & 0.19 & 0.21 & 0.22 & 0.22 & 0.32 \\
     \hline
     \multirow{8}{*}{\rotatebox{90}{OR-CP}} & 0.1 & 1.94 & 2.01 & 1.81 & 2.15 & 2.22 & 2.47 & 2.6 & 2.13 & 11.04 \\
    & 0.2 & 1.57 & 1.6 & 1.32 & 1.01 & 0.96 & 1.23 & 1.15 & 1.73 & 45.98[9] \\
    & 0.3 & 0.7 & 0.76 & 0.78 & 0.83 & 0.9 & 0.96 & 1.14 & 4.07 & 48.42[4] \\
    & 0.4 & 1.04 & 1.08 & 1.11 & 1.13 & 1.33 & 1.47 & 2.07 & 3.88 & 474.78[5] \\
    & 0.5 & 0.41 & 0.41 & 0.45 & 0.47 & 0.5 & 0.56 & 1.6 & 9.06 & 434.35[5] \\
    & 0.6 & 0.33 & 0.33 & 0.36 & 0.4 & 0.4 & 0.7 & 1.01 & 19.24 & 76.53[1] \\
    & 0.7 & 0.26 & 0.27 & 0.28 & 0.3 & 0.32 & 0.4 & 8.62 & 82.19[6] & 163.76[6] \\
    & 0.8 & 0.23 & 0.23 & 0.24 & 0.25 & 0.29 & 1.75 & 1.64 & 9.4 & 80.41[8] \\
    \hline
    \multirow{8}{*}{\rotatebox{90}{\clingo}} &  0.1 & TO  & TO & TO  & TO & TO & TO  & TO & TO & TO \\
    & 0.2  & TO & TO & TO & TO & TO & TO  & TO & TO & TO  \\
    & 0.3 & TO &  TO  & TO  & TO & TO & TO & TO &  TO & TO  \\
    & 0.4  & 1909.7[1] & TO & 877.16[2] & TO & TO & TO &  TO  & TO & TO\\
    & 0.5 & 1305.27[2] & 592.94[3] & 297.8[2] & 1658.6[1]  & 953.01[1] & TO &  1182.33[2] & 1437.34[3] & TO  \\
    & 0.6 & 835.24[7] & 656.42 & 1159.53[4] & 910.1[4] & 1093.83[4] & 1048.79[8] & 722.22[7] & 758.89[3] & TO \\
    & 0.7 & 82.61 & 70.12 &  240.72 & 211.46 &  115.87 & 245.94  & 906.8[8] &  488.83[4] & 896.77[2] \\
    & 0.8 & 6.98 & 9.33 & 19.6 & 13.45 & 17.54 & 70.72 & 34.5 & 211.64 & 356.36[7] \\
        \hline \hline
    \end{tabular}}}
    \\
    \footnotesize{TO: Timeout (over 2000 seconds)}
\end{minipage}
   \vspace{-\baselineskip}
\end{table}

    \begin{table}[t]
     \caption{Egalitarian \smti: Average CPU-Times (in seconds) for $n=50$.}
    \label{tab:optimizationE}
       \vspace{-.3\baselineskip}
    \begin{minipage}{\textwidth}
    {\footnotesize    \centering
    \begin{tabular}{p{6ex}p{2ex}p{3ex}p{3ex}p{3ex}p{3ex}p{4ex}p{8ex}p{9ex}p{9ex}p{9ex}}
    \hline \hline
       &  &  \multicolumn{9}{c}{$p2$} \\
             \cline{3-11}\\
    Solver & $p1$ & 0.1 & 0.2 & 0.3 & 0.4 & 0.5 & 0.6 & 0.7 & 0.8 & 0.9 \\
    \hline
    \multirow{8}{*}{Gurobi}
     & 0.1 & 0.42 & 0.44 & 0.46 & 0.46 & 0.48 & 0.5 & 0.52 & 0.59 & 0.66 \\
     & 0.2 & 0.34 & 0.35 & 0.36 & 0.39 & 0.4 & 0.42 & 0.49 & 0.57 & 0.65 \\
     & 0.3 & 0.27 & 0.29 & 0.29 & 0.32 & 0.34 & 0.35 & 0.39 & 0.48 & 0.85 \\
     & 0.4 & 0.21 & 0.22 & 0.22 & 0.24 & 0.27 & 0.28 & 0.37 & 0.43 & 0.48 \\
     & 0.5 & 0.17 & 0.16 & 0.18 & 0.19 & 0.19 & 0.23 & 0.25 & 0.32 & 0.59 \\
     & 0.6 & 0.12 & 0.12 & 0.14 & 0.15 & 0.15 & 0.18 & 0.2 & 0.27 & 0.32 \\
     & 0.7 & 0.09 & 0.1 & 0.1 & 0.11 & 0.11 & 0.12 & 0.15 & 0.17 & 0.31 \\
     & 0.8 & 0.07 & 0.08 & 0.09 & 0.09 & 0.09 & 0.1 & 0.1 & 0.12 & 0.16 \\
    \hline
    \multirow{8}{*}{OR-MIP}
    & 0.1 & 0.7 & 0.72 & 0.73 & 0.75 & 0.76 & 0.79 & 0.83 & 0.99 & 1.3 \\
    & 0.2 & 0.56 & 0.58 & 0.59 & 0.61 & 0.62 & 0.67 & 0.7 & 1.01 & 1.28 \\
    & 0.3 & 0.45 & 0.47 & 0.48 & 0.51 & 0.52 & 0.53 & 0.63 & 0.98 & 1.9 \\
    & 0.4 & 0.35 & 0.36 & 0.37 & 0.39 & 0.44 & 0.47 & 0.66 & 0.86 & 1.14 \\
    & 0.5 & 0.28 & 0.27 & 0.29 & 0.32 & 0.32 & 0.37 & 0.45 & 0.75 & 1.28 \\
    & 0.6 & 0.21 & 0.21 & 0.22 & 0.24 & 0.25 & 0.35 & 0.37 & 0.58 & 0.93 \\
    & 0.7 & 0.15 & 0.15 & 0.16 & 0.18 & 0.19 & 0.2 & 0.28 & 0.42 & 0.63 \\
    & 0.8 & 0.11 & 0.11 & 0.12 & 0.12 & 0.14 & 0.16 & 0.17 & 0.24 & 0.39 \\
    \hline
     \multirow{8}{*}{OR-CP} & 0.1 & 1.01 & 1.04 & 1.07 & 1.38 & 2.13 & 96.34 & 417.34[5] & TO & TO\\
    & 0.2 & 1.43 & 1.51 & 1.19 & 1.85 & 3.71 & 196.42 & 391.73[6] & TO & TO\\
    & 0.3 & 1.1 & 1.19 & 1.33 & 1.78 & 11.76 & 60.11[8] & 561.85[4] & TO & TO\\
    & 0.4 & 0.9 & 1.0 & 1.01 & 1.19 & 9.06 & 108.64 & 629.32[4] & TO & TO\\
    & 0.5 & 0.79 & 0.77 & 0.81 & 1.14 & 1.97 & 147.54 & 213.31[3] & TO & TO \\
    & 0.6 & 0.6 & 0.61 & 0.65 & 1.84 & 2.25 & 131.32 & 82.9[1] & TO & TO \\
    & 0.7 & 0.45 & 0.45 & 0.49 & 1.19 & 1.21 & 26.87 & 951.98[1] & TO & TO\\
    & 0.8 & 0.35 & 0.36 & 0.36 & 0.44 & 1.48 & 55.2[8] & 477.63[3] & TO & TO\\
    \hline
    \multirow{8}{*}{\clingo} & 0.1 & 5.33 & 5.23 & 5.34 & 5.58 & 6.09 & 32.46 & 786.3[8] & 714.67[1] & TO \\
    & 0.2 & 4.03 & 4.07 & 4.14 & 4.3 & 5.36 & 48.34 & 309.89[8] & TO & TO \\
    & 0.3 & 2.48 & 2.17 & 2.52 & 2.81 & 4.5 & 220.81 &  152.69[5] & TO & TO  \\
    & 0.4 & 1.46 & 22.77 & 1.49 & 1.56 & 3.65 & 19.33 &  365.44[6] & TO & TO \\
    & 0.5 & 0.82 & 0.79 & 0.82 & 0.91 & 1.1 & 29.54 &  78.39[3] & TO & TO  \\
    & 0.6 & 0.42 & 0.39 & 0.43 & 0.75 & 0.89 & 34.48 & 329.08[3] & TO & TO \\
    & 0.7 & 0.2 & 0.2 & 0.21 & 0.44 & 0.49 & 5.93 & 564.82[1] & TO & TO \\
    & 0.8 & 0.09 & 0.09 & 0.09 & 0.14 & 0.76 &  47.57[8] & 430.35[3] & TO & TO \\
    \hline \hline
    \end{tabular}}

    \footnotesize{TO: Timeout (over 2000 seconds)}
\end{minipage}
   \vspace{-\baselineskip}
\end{table}

For Sex-Equal \smti and Egalitarian \smti, we have compared the ILP approaches using Gurobi and Google-OR Tools MIP, the CP approach using Google-OR Tools CP, and the ASP approach using \clingo. The results are shown in Tables~\ref{tab:optimizationS} and~\ref{tab:optimizationE} for $n=50$.

For Sex-Equal \smti and Egalitarian \smti, we observe that the average CPU times are larger for the CP and ASP approaches when compared with the Max Cardinality \smti experimental results (Table~\ref{tab:maxCard50}).  In the tables, the numbers in square brackets denote how many of the 10 instances could be solved. Note that there is a larger number of instances that could not be solved with these approaches within the given time threshold. The ILP approaches, on the other hand, perform better.

It is interesting to observe that \clingo has better results for Max Cardinality \smti, compared to Egalitarian \smti and Sex-Equal \smti. This may be related to the objective functions: Egalitarian \smti and Sex-Equal \smti aims to minimize the sum of costs that are calculated by addition or subtraction of some nonnegative numbers whereas Max Cardinality \smti aims to minimize the sum of 1's. The large CPU times for the ASP approach for Sex-Equal \smti could also be due to the use of aggregates in weak constraints.

\section{Conclusion}\label{sec:conc}

We have conducted an empirical study to compare different approaches to solve hard instances of \smti: Max Cardinality \smti, Sex-Equal \smti, and Egalitarian \smti.  For that, we have introduced formulations of these problems in Answer Set Programming utilizing weak constraints, and implemented them for \clingo. We have adapted an existing Integer Linear Programming model of Max Cardinality \smti, for other optimization variants of \smti, and implemented them for Gurobi and Google-OR Tools (MIP and CP). We have also implemented two different existing local search algorithms to solve Max Cardinality \smti, with slight adaptations. We have compared these approaches empirically over randomly generated instances.

We have also performed experiments to compare Answer Set Programming with Propositional Satisfiability, over \smti instances. For the latter, we have utilized \cmodels with the SAT solver \zchaff, and SAT-E implementation with the SAT solver \lingeling.

There are several important discussions. First of all, modeling is an important step in all these problem-solving methods. For that reason, we have utilized the existing and empirically evaluated models in the literature, for SAT, ILP and local search. We have come up with our own ASP formulation for \smti variants after trying different formulations and considering elaboration tolerance, as the existing ASP formulations of the stable marriage problems used as benchmarks in the ASP competitions address \smt problem (a tractable variant of \sm~\cite{irving1994}).

Although Google-OR Tools provide a CP solver that takes as input (almost the same) ILP formulations of \smti problems, we think it will be interesting to introduce CP formulations of these problems (e.g., in the spirit of~\cite{gent2002empirical}) and experiment with some other CP solvers. Similarly, although \cmodels provides a SAT-based method to solve \smti instances, it is hinted by our experiments with the SAT encodings of \smti by \citeN{Drummond15sat} that it will be worthwhile to compare our results with different SAT encodings of \smti (e.g., in the spirit of \cite{Gent02satencodings}) and to extend SAT-E to solve optimization variants. Furthermore, more sophisticated ILP models for Max Cardinality \smti (e.g., in the spirit of \cite{DelormeGGKMP19}) could be considered. These studies are part of our future work.

Regarding the experimental results for optimization variants, it is interesting to observe that the declarative methods (ASP, ILP, CP) are more promising compared to the local search algorithms as the problems get harder with more ties and incompleteness.
It is also interesting to observe that the ILP approaches are in general better than the ASP and CP approaches in terms of computation time.

We believe that comparing different (but closely related) methods to solve hard problems is valuable to better understand their strengths~\cite{cayli2007,coban2008,DovierFP09,fidan20}. Our study of the hard variants of \smti problems contributes to this line of research, not only by providing models and implementations but also by providing benchmarks for future studies.


\paragraph{\bf Acknowledgments.} We would like to thank Ian Gent, David Manlove, Andrew Perrault, William Pettersson and Patrick Prosser for useful discussions and suggestions, and sharing their software with us. We would also like to thank anonymous reviewers for their valuable comments.

\clearpage
\appendix
\section{Algorithms}
\begin{algorithm}[H]
\caption{LTIU}
\textbf{Input: } An \smti instance of size $n$, step limit $S$\\
\textbf{Output: } A matching $\mu$

\begin{algorithmic}\label{alg:ltiu}
\algrule
\STATE $\mu \gets \text{a randomly generated matching}$
\STATE $\mu_{best} \gets \mu$
\STATE $step \gets 0$
\WHILE{$step < S$}
\IF{$\mu$ is a perfect matching}
\STATE $\mu_{best} \gets \mu$
\STATE break
\ENDIF
\STATE $bps \gets \{(m,w) | (m,w)$ is an undominated blocking pair$\}$
\IF{$bps$ is empty}
\IF{$\mu$ is better than $\mu_{best}$}
\STATE $\mu_{best} \gets \mu$
\ENDIF
\STATE $\mu \gets \text{a randomly generated matching}$
\ELSE
\STATE $neighbors \gets \{$removing $(m,w)$ from $ \mu | (m,w) \in bps \}$
\STATE pick a random number $r$ in $(0,1)$
\IF{$r < p$}
\STATE $\mu \gets$ a random matching  in $neighbors$
\ELSE
\STATE $eval_\mu \gets$ number of singles $+$ number of blocking pairs in $\mu$
\FOR{$n \in neighbors$}
\STATE $eval_n \gets$ number of singles $+$ number of blocking pairs in $n$
\ENDFOR
\IF{$eval_\mu > argmin(eval)$}
\STATE  $\mu \gets $ a random neighbor $n$ such that $eval_n = argmin(eval)$
\ELSE
\STATE  $\mu \gets $ a random matching in $neighbors$
\ENDIF
\ENDIF
\ENDIF
\STATE  $step = step + 1$
\ENDWHILE
\RETURN $\mu_{best}$
\end{algorithmic}
\end{algorithm}

\begin{algorithm}[h]
\caption{Adapted Genetic Algorithm}
\textbf{Input: } An \smti instance $I$, probability for crossover $p$, probability for  mutation $m$, number of evolution rounds $n$ and size of population $S$

\textbf{Output: } A matching $\mu$
\algrule
\begin{algorithmic}\label{alg:ga}
\STATE $population \gets \emptyset$
\FOR{$i \gets 1$ \textbf{to} $S$}
\STATE $\mu_0 = $ obtain a matching by applying DA on $I$
\STATE add $\mu_0$ to $population$
\ENDFOR
\FOR{$i \gets 1$ \textbf{to} $evolution\ rounds$}
\STATE $temporary\ population \gets population$
\STATE select pairs of solutions based on crossover probability and perform cycle crossover operation
\FOR{$solution \in temporary\ population$}
\STATE select a random number $r$ in $(0,1)$
\IF{$r < m$}
\STATE mutate $solution$
\ENDIF
\ENDFOR
\STATE $population \gets$ best solutions from temporary population
\ENDFOR
\RETURN $\mu \gets$ best solution in population
\end{algorithmic}
\end{algorithm}
\clearpage

\section{Further Experimental Results}
\begin{table}[H]
\caption{Average number of atoms for varying $p1$ and $p2$ values with $n=50$}
     \label{tab:smti50-atoms}
   \begin{minipage}{\linewidth}
    \begin{tabular}{p{2ex}p{2ex}p{5ex}p{5ex}p{5ex}p{5ex}p{5ex}p{5ex}p{5ex}p{5ex}p{7ex}}
                   \hline \hline
    \multirow{3}{*}{\rotatebox{90}{Solver}} &  &  \multicolumn{9}{c}{$p2$} \\
            \cline{3-11}\\
      & $p1$ & 0.1 & 0.2 & 0.3 & 0.4 & 0.5 & 0.6 & 0.7 & 0.8 & 0.9 \\
      \hline
      \multirow{8}{*}{\rotatebox{90}{\cmodels}}   & 0.1 & 113425 & 112634 & 112320 & 110987 & 111095 & 109762 & 108012 & 104989 & 96243\\
 & 0.2 & 89963 & 90285 & 90230 & 89101 & 88298 & 87244 & 85941 & 83127 & 7602\\
 & 0.3 & 71035 & 70487 & 70567 & 70404 & 69397 & 69143& 67309 & 64401 & 57927\\
 & 0.4 & 53118 & 52825 & 51973 & 52700 & 51603 & 51357 & 49781 & 48818 & 43102\\
 & 0.5 & 38439 & 37674 & 37980 & 37733 & 37144 & 37668 & 34827 & 33615 & 30541\\
 & 0.6 & 26444 & 25382 & 26313 & 25192 & 24580 & 24654 & 23674 & 22664 & 19953\\
 & 0.7 & 15863 & 15638 & 15549 & 15338 & 15134 & 15239 & 14381 & 14004 & 12168\\
 & 0.8 & 8078 & 8315 & 8229 & 7929 & 8014 & 7647 & 7400 & 7004 & 6373\\
       \hline
      \multirow{8}{*}{\rotatebox{90}{SAT-E}}
        & 0.1 & 9022 & 8997 & 8984 & 8931 & 8926 & 8849 & 8758 & 8572 & 7963\\
 & 0.2 & 7969 & 7991 & 7988 & 7946 & 7901 & 7858 & 7783 & 7573 & 7086\\
 & 0.3 & 7013 & 6995 & 7006 & 6994 & 6941 & 6951 & 6823 & 6658 & 6156\\
 & 0.4 & 5986 & 5972 & 5931 & 5958 & 5911 & 5891 & 5791 & 5745 & 5234\\
 & 0.5 & 5001 & 4952 & 4982 & 4973 & 4933 & 4975 & 4787 & 4666 & 4317\\
 & 0.6 & 4049 & 3960 & 4052 & 3966 & 3914 & 3935 & 3856 & 3782 & 3425\\
 & 0.7 & 3004 & 2990 & 2979 & 2974 & 2962 & 2960 & 2899 & 2850 & 2635\\
 & 0.8 & 1975 & 2024 & 2013 & 1970 & 1991 & 1957 & 1920 & 1861 & 1716\\
                \hline \hline

    \end{tabular}
    \end{minipage}
\end{table}

\begin{table}[h]
    \caption{Average number of clauses for varying $p1$ and $p2$ values with $n=50$}
     \label{tab:smti50-clauses}
   \begin{minipage}{\linewidth}
    {\centering
    \begin{tabular}{p{2ex}p{2ex}p{6ex}p{6ex}p{6ex}p{6ex}p{6ex}p{6ex}p{6ex}p{6ex}p{6ex}}
               \hline \hline
      \multirow{3}{*}{\rotatebox{90}{Solver}}  &  &  \multicolumn{9}{c}{$p2$} \\
            \cline{3-11}\\
     & $p1$ & 0.1 & 0.2 & 0.3 & 0.4 & 0.5 & 0.6 & 0.7 & 0.8 & 0.9 \\
      \hline
      \multirow{8}{*}{\rotatebox{90}{\cmodels}}   & 0.1 & 1296263 & 1279570 & 1276494 & 1254057 & 1255637 & 1229722 & 1191007 & 1129910 & 962283\\
 & 0.2 & 937668 & 937568 & 935593 & 919152 & 905123 & 887740 & 864279 & 813669 & 681242\\
 & 0.3 & 670565 & 659872 & 660876 & 659103 & 641055 & 637263 & 609766 & 562206 & 464503\\
 & 0.4 & 443675 & 439157 & 428319 & 435531 & 421185 & 417162 & 395655 & 377376 & 297961\\
 & 0.5 & 280194 & 271381 & 274901 & 270723 & 263916 & 268156 & 237253 & 221306 & 181749\\
 & 0.6 & 164589 & 154443 & 162863 & 152339 & 147149 & 146310 & 137174 & 124952 & 98137\\
 & 0.7 & 79703 & 77568 & 77378 & 75205 & 73489 & 73914 & 67096 & 62831 & 48854\\
 & 0.8 & 31065 & 31959 & 31535 & 29783 & 29984 & 27705 & 26257 & 23537 & 19860\\
       \hline
      \multirow{8}{*}{\rotatebox{90}{SAT-E}}
         & 0.1 & 124392 & 123861 & 123903 & 122977 & 123803 & 123371 & 123251 & 123285 & 121668\\
 & 0.2 & 99573 & 100199 & 100467 & 99706 & 99439 & 99280 & 99371 & 98668 & 98449\\
 & 0.3 & 79429 & 79100 & 79484 & 79675 & 79133 & 79664 & 78818 & 78203 & 76368\\
 & 0.4 & 60204 & 60081 & 59432 & 60498 & 59797 & 60148 & 59490 & 60299 & 58423\\
 & 0.5 & 44271 & 43583 & 44159 & 44155 & 43889 & 44973 & 42697 & 42663 & 42386\\
 & 0.6 & 31061 & 29994 & 31231 & 30205 & 29783 & 30322 & 29808 & 29820 & 28736\\
 & 0.7 & 19138 & 18983 & 18981 & 18931 & 18908 & 19254 & 18798 & 19014 & 17933\\
 & 0.8 & 10040 & 10403 & 10369 & 10090 & 10306 & 10086 & 9972 & 9849 & 9523\\
\hline \hline
    \end{tabular}}
    \end{minipage}
    \vspace{-\baselineskip}
\end{table}

\clearpage
\bibliographystyle{acmtrans}
\bibliography{reference}

\label{lastpage}

\end{document}